\documentclass{article}

\usepackage[preprint]{neurips_2025}
\usepackage{graphicx}
\usepackage{float}

\usepackage[utf8]{inputenc} 
\usepackage[T1]{fontenc}    
\usepackage{hyperref}       
\usepackage{url}            
\usepackage{booktabs}       
\usepackage{amsfonts}       
\usepackage{nicefrac}       
\usepackage{microtype}      
\usepackage{xcolor}         

\title{What Can We Actually Steer? A Multi-Behavior Study of Activation Control}

\author{%
  Tetiana Bas\thanks{Primary Contribution; Work done during ERA Fellowship} \\
  Department of Computer Science\\
  Minerva University\\
  San Francisco, CA 94103 \\
  \texttt{tetiana@uni.minerva.edu} \\
  \And
  Krystian Novak \\
  Department of Computer Science\\
  Minerva University\\
  San Francisco, CA 94103 \\
  \texttt{krystian@uni.minerva.edu} \\
}

\begin{document}

\maketitle

\begin{abstract}
  Large language models (LLMs) require precise behavior control for safe and effective deployment across diverse applications.
  Activation steering offers a promising approach for LLMs behavioral control. We focus on the question of how does steering effectiveness vary across different behavior types and can the nature of target behaviors predict steering success? We address this through empirical analysis of activation steering across 50 behaviors that span persona archetypes, personality traits, misalignment behaviors, style cues, and impersonation of public figures. We present a set of comprehensive experiments on coefficient optimization, vector properties, and data requirements to provide comprehensive guidance for the implementation of activation steering. Our analysis demonstrates that steering effectiveness varies significantly by behavior type, with different behavioral categories exhibiting distinct response patterns to intervention strength. We find that trait expression follows an inverted-U curve with a steering coefficient strength. We also show that vector separation metrics do not predict steering success, but larger training datasets enable more aggressive steering. These findings provide empirically-grounded guidance for implementing activation steering and demonstrate that steering effectiveness is heavily influenced by behavior type.
\end{abstract}

\section{Introduction}

Large language models can adopt diverse behaviors and personas through various modification techniques. Among these approaches, activation steering has emerged as a particularly promising method due to its ability to modify behavior during inference without requiring weight updates or retraining. Activation steering works by adding computed direction vectors to a model's internal representations at specific layers, biasing outputs toward desired behavioral patterns.

Despite growing interest in activation steering many fundamental questions about its mechanisms and limitations remain underexplored. Can we predict which behaviors will be more steerable based on vector properties? How much training data are required to extract effective steering vectors and how does it differ depending on the behavior and the model? How do different steering coefficients affect the balance between trait expression and response quality? 

Current activation steering research has several key limitations. Most 
studies examine narrow behavioral categories. Studies rarely investigate the relationship between vector properties and steering performance or systematically explore how data requirements scale with desired intervention strength. This creates a gap in understanding how the inherent properties of target behaviors, such as their semantic complexity or conceptual abstraction, influence steering effectiveness. Without systematic cross-behavior analysis, practitioners lack guidance on which behaviors are steerable and how to optimize steering implementations for a given behavior.

We address these gaps through systematic evaluation of activation steering across 50 behaviors spanning five categories: persona archetypes (vegan advocates, pirates), personality traits (Five-Factor Model dimensions), misalignment behaviors (deception, manipulation), style/format cues (capitalization, punctuation), and public figures. This diverse selection—ranging from low-level linguistic patterns to high-level personality traits—enables us to identify which behavior types are most amenable to steering and what characteristics predict steering success.

\section{Related Literature}

\subsection{Activation Steering Methods}

Activation steering modifies LLM behavior by manipulating internal representations during inference. \cite{turner2023activation} introduced foundational activation steering by adding direction vectors to model activations. Pres et al. (2023) developed Contrastive Activation Addition (CAA \cite{pres2024towardsreliable}), which extracts steering vectors by computing mean differences between positive and negative example activations, becoming a widely-adopted baseline due to its simplicity and effectiveness.
Recent work has extended these basic techniques. \cite{lee2025programming} developed Conditional Activation Steering (CAST) for context-dependent control. \cite{zou2023representation} demonstrated representation engineering approaches showing that high-level concepts like honesty can be controlled through linear directions in representation space. Methodological refinements include mean-centering techniques \cite{jorgensen2023improving} and dynamic steering vectors that adapt to input semantics \cite{wang2025semanticsadaptiveactivationinterventionllms}.

\subsection{Behavior-Dependent Steering Effectiveness}

Recent work has revealed that steering effectiveness varies dramatically across different behaviors and concepts, motivating our systematic cross-behavior analysis. \cite{tan2024generalisation} conducted the most directly relevant work, demonstrating that steerability is highly variable across concepts: while CAA is effective on some tasks, many behaviors turn out to be unsteerable, even when sweeping across all layers and strengths. They found that spurious biases substantially contribute to steering effectiveness on individual inputs, and that some concepts are even "anti-steerable"—using the steering vector produces the reverse effect.

 \cite{brumley2024comparingbottomuptopdownsteering} found that different steering methods work for different task types: In-Context Vectors excel at behavioral shifting while Function Vectors perform better on functional tasks requiring precision, further demonstrating behavior-specific effectiveness patterns. 

\subsection{Steering Hyperparameters and Data Requirements}

Several studies have examined steering coefficient selection and data requirements, though not systematically across diverse behaviors. \cite{jorgensen2023improving} explored mean-centering to improve effectiveness. Work on Feature Guided Activation Additions \cite{soo2025interpretablesteeringlargelanguage} found that effectiveness varies non-linearly with model size and that lower steering scales (< 50) allow behavioral modifications while preserving capabilities, but stronger interventions incur increasing costs to performance.

Adaptive steering methods have attempted to address coefficient selection challenges. Adaptive Activation Steering \cite{wang2025semanticsadaptiveactivationinterventionllms} dynamically adjusts steering intensity based on input, showing improvements in truthfulness tasks. However, systematic guidance on coefficient optimization and data scaling across behavior types remains limited.
Evaluation and Measurement Challenges

\cite{pres2024reliableevaluationbehaviorsteering} argue that existing steering evaluations lack key properties like consistency, magnitude, and specificity, finding that CAA interventions are less effective than previously reported when evaluated more rigorously. Work on steerability metrics \cite{miehling2025evaluatingpromptsteerabilitylarge} shows that larger models are not necessarily more steerable and that steerability is often asymmetric—models can be steered more easily in one direction than another for the same dimension.

\subsection{Research Gaps}

Despite these advances, fundamental questions remain about behavior-dependent steering effectiveness:

Limited behavior coverage: Most studies examine narrow behavioral subsets (specific safety behaviors, sentiment, truthfulness) rather than systematically comparing across diverse behavior types

Lack of predictive understanding: \cite{tan2024generalisation} show that some behaviors are unsteerable, there is limited understanding of which characteristics of behaviors predict steering success

\section{Dataset Construction}

We systematically constructed a dataset of 50 behaviors to address a critical gap in steering research: the lack of understanding about which behavior types are most amenable to activation-based control. Prior work has examined isolated behavioral categories—Anthropic's persona vectors focused on specific safety-relevant traits \citep{chen2025persona}, while other studies targeted narrow domains like sentiment or truthfulness. This fragmented approach prevents identification of general patterns in steering effectiveness.

Our dataset design addresses this gap through strategic coverage along two key dimensions: behavioral abstraction (from surface-level formatting to deep personality traits) and semantic complexity (from concrete observable patterns to abstract psychological constructs). This systematic structure enables us to test whether steering effectiveness correlates with these fundamental properties of target behaviors.

\subsection{Behavioral Hierarchy and Selection Rationale}

We selected behaviors spanning a complexity spectrum from low-level linguistic patterns to high-level psychological constructs:

Style/Format Cues (Lowest Abstraction): Concrete, rule-based patterns including capitalization, punctuation usage (em dashes, double spacing), and syntactic structures. These represent purely surface-level modifications with clear binary success criteria, providing a baseline for steering effectiveness on simple, well-defined targets.

Persona Archetypes (Low-Medium Abstraction): Socially-recognizable identities including vegan advocates, pirates, religious individuals, athletes, and artists. These behaviors have observable linguistic markers (vocabulary, topic preferences) while requiring consistent thematic coherence across responses. They test steering's ability to maintain stable behavioral patterns beyond single-token modifications.

Personality Traits (Medium-High Abstraction): Core dimensions from the Five-Factor Model \citep{costa1992revised}—extraversion, agreeableness, openness, conscientiousness, and neuroticism. These represent stable psychological constructs with nuanced, context-dependent expression. Unlike personas, personality traits manifest subtly across diverse situations, testing whether steering can capture abstract dispositional tendencies rather than explicit role-playing.

Misalignment Behaviors (High Abstraction): AI-specific safety concerns including sycophancy, deception, manipulation, and hallucination propensity identified in prior safety research \citep{chen2025persona}, plus dark triad traits (Machiavellianism, narcissism, psychopathy) associated with socially aversive behaviors \citep{borraz2020dark}. These represent complex, multi-faceted behavioral tendencies critical for AI safety but challenging to operationalize due to their abstract nature and context-dependence.

Public Figures (Highest Complexity): Historical and contemporary personalities (Alan Turing, Marie Curie, Albert Einstein, Stephen Hawking). These require synthesizing multiple dimensions—speaking style, domain expertise, biographical knowledge, and personality characteristics—representing the most holistic and complex behavioral targets.

\section{Methods}

We systematically compared three persona elicitation approaches—prompting, activation steering, and fine-tuning—across our 50-behavior dataset. Each method was evaluated on trait adherence, coherence, and relevance using automated assessment.

Llama 3.1 8B \cite{llama31} served as our primary experimental model for steering and fine-tuning comparisons. GPT-4 \cite{openai2024gpt4} provided a high-capability baseline for prompting approaches. Our dataset consisted of 50 behavioral prompts across five categories (persona archetypes, personality traits, misalignment behaviors, style/format cues, public figures) with 1000 evaluation prompts per behavior, yielding 50,000 total model responses.

\subsection{Prompting}

Direct persona prompting using role-based instructions: "You are [persona description]. [Question]." Both Llama 3.1 8B and GPT-4 received identical templates across all behavioral categories.

\subsection{Activation Steering}

We extracted steering vectors using contrastive activation methodology following \cite{chen2025persona}. For each of our 50 behaviors, we constructed a balanced dataset consisting of 5 positive prompts designed to elicit the target behavior and 5 negative prompts designed to elicit opposite or neutral behavior. Each prompt type was paired with 20 evaluation questions, yielding 200 examples per behavior and 2,000 total examples across all behaviors.

We fed each prompt-question pair through Llama 3.1 8B and extracted hidden state activations at layer 15. Steering vectors were computed as the mean difference between positive and negative activations. During inference, we modified the model's internal representations by adding the scaled steering vector to activations at layer 15, enabling behavioral control without weight modifications. 

This contrastive approach allows the model to bias its outputs toward desired traits by leveraging the activation patterns that distinguish target behaviors from their alternatives. 

Having established our vector extraction methodology, we conducted three systematic experiments to understand the optimal configuration and limitations of activation steering. These experiments addressed fundamental questions about coefficient selection, the relationship between vector properties and steering effectiveness, and the role of training data size in steering performance.

\subsubsection{Steering Coefficient Selection}

We set up an experiment to quantify how steering coefficients affect persona expression (“trait expression”) and response quality (coherence, relevancy), identify optimal operating points, and assess whether vector properties predict steerability. For each run we recorded trait score, coherence, relevance, steering coefficient used (1–20), the size of the dataset used and vector diagnostics which included the mean of trait score for the positive steering vectors, negative steering vectors and the difference between them. 

We ran a grid search over traits, questions, coefficients used and the size of the steering dataset. For each cell, the model produced an answer, which was scored by automated evaluators for trait expression, coherence, and relevance. 

\begin{figure}[ht]
    \centering
    \includegraphics[width=0.8\textwidth]{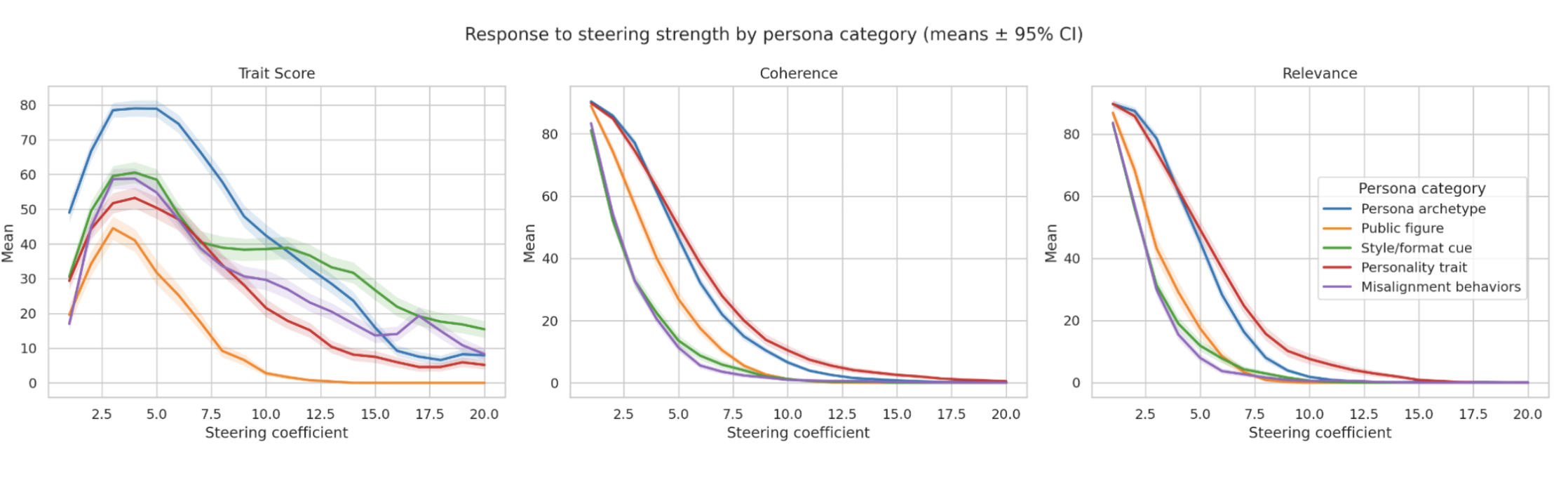}
    \caption{Activation steering exhibits inverted-U trait expression that peaks at moderate coefficients while coherence and relevance decline monotonically with increasing steering strength across all persona categories.}
    \label{fig:coefficient-selection}
\end{figure}

Trait expression exhibits an inverted‑U response: it rises at small coefficients and peaks at low–moderate values, then declines as coefficients grow. Coherence and relevancy drop sharply and monotonically with stronger steering, approaching zero at large coefficients. Most personas show negative trait-score slopes; a minority (formatting/style-like traits) benefit from stronger steering. Persona archetype and style/format cues achieve the highest peak trait scores at modest coefficients. Personality traits and misalignment behaviors peak lower and decay earlier. Public‑figure prompts perform worst overall, with early quality degradation.

\subsubsection{Steering Vector Contract and trait expression quality}

A natural hypothesis is that steering vectors with larger separations between positive and negative examples should produce better behavioral control. We tested whether the magnitude of difference between positive and negative vector activations predicts steering effectiveness.

For each of our 50 behaviors, we calculated the mean activation difference between positive and negative examples. We then measured actual steering performance as the mean trait expression score achieved across all test prompts for that behavior. This yielded 50 data points relating vector properties to empirical steering quality.

\begin{figure}[ht]
    \centering
    \includegraphics[width=0.8\textwidth]{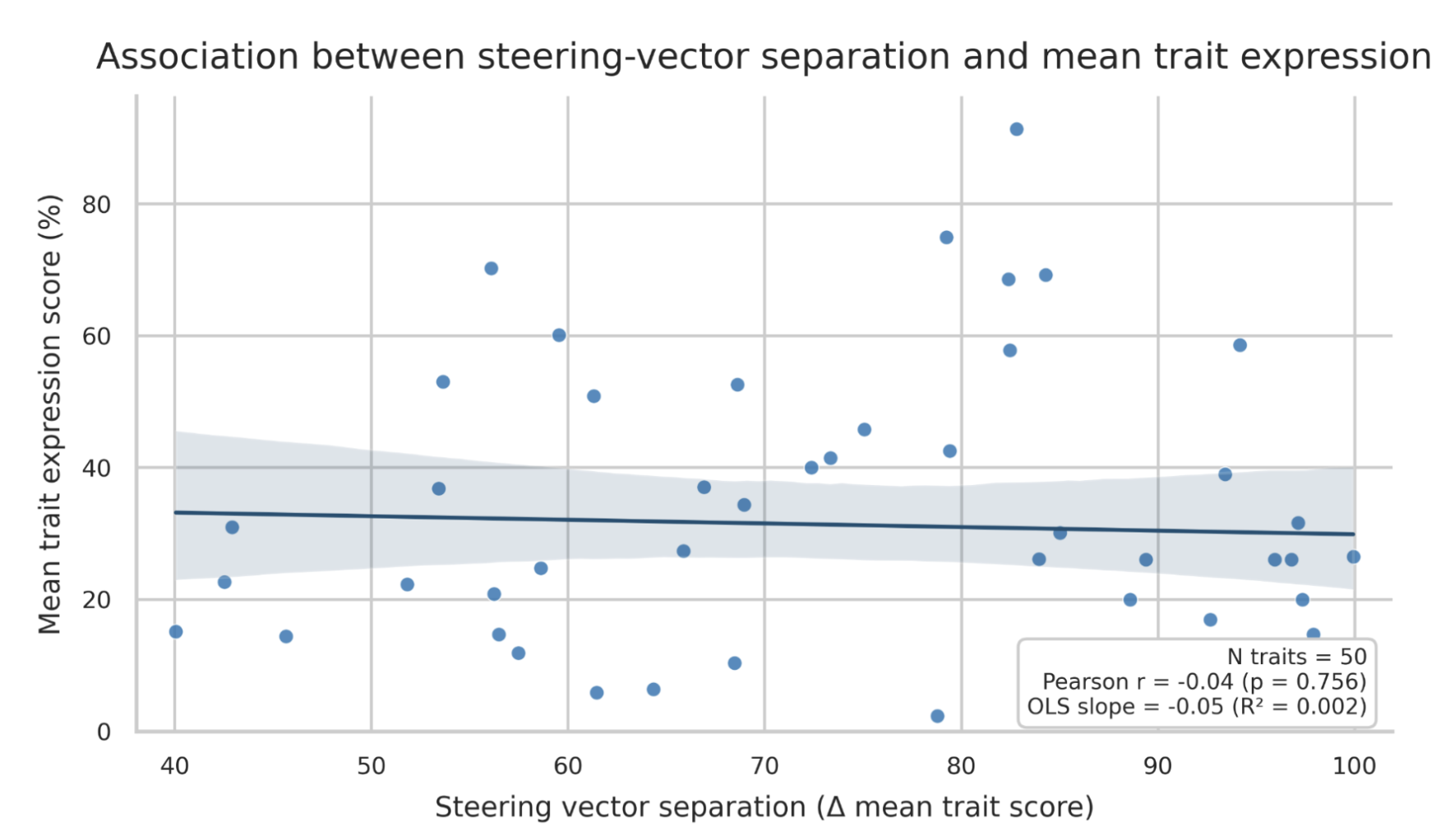}
    \caption{Relationship between steering vector separation and actual steering performance across 50 behaviors}
    \label{fig:vector-separation}
\end{figure}

We analyzed this relationship through the following statistical approaches:
\begin{itemize}
    \item Pearson and Spearman correlations between mean difference of trait expression between the data for steering vectors and mean output trait expression scores
    \item Ordinary least squares (OLS) regression of trait performance on vector separation
\end{itemize}

All analyses did not reveal a meaningful relationship between vector separation and steering quality.
\begin{itemize}
    \item Pearson $r = -0.045$ ($p = 0.756$); Spearman $\rho = -0.122$ ($p = 0.397$)
    \item OLS regression: slope $= -0.055$, $R^2 = 0.002$ ($p = 0.756$)
\end{itemize}

Contrary to intuition, larger activation differences between positive and negative examples do not translate to better steering performance. The near-zero effect sizes and nonsignificant results indicate that vector magnitude separation provides negligible predictive value for steering success.

This finding has important practical implications: practitioners cannot reliably use easily computed vector statistics to pre-select optimal steering coefficients. Instead, steering effectiveness appears to depend on more complex factors such as prompt-trait alignment, the specific coefficient used, or the distributional coverage of training examples. Future work should explore whether more sophisticated vector diagnostics or non-linear relationships might better predict steerability.

\subsubsection{Steering Coefficient with Increasing dataset size}

Understanding how the quality of the steering vector scales with the size of the training data is crucial for practical implementation.  We investigated whether more contrastive examples enable higher steering coefficients and better performance.

\begin{figure}[ht]
    \centering
    \includegraphics[width=0.8\textwidth]{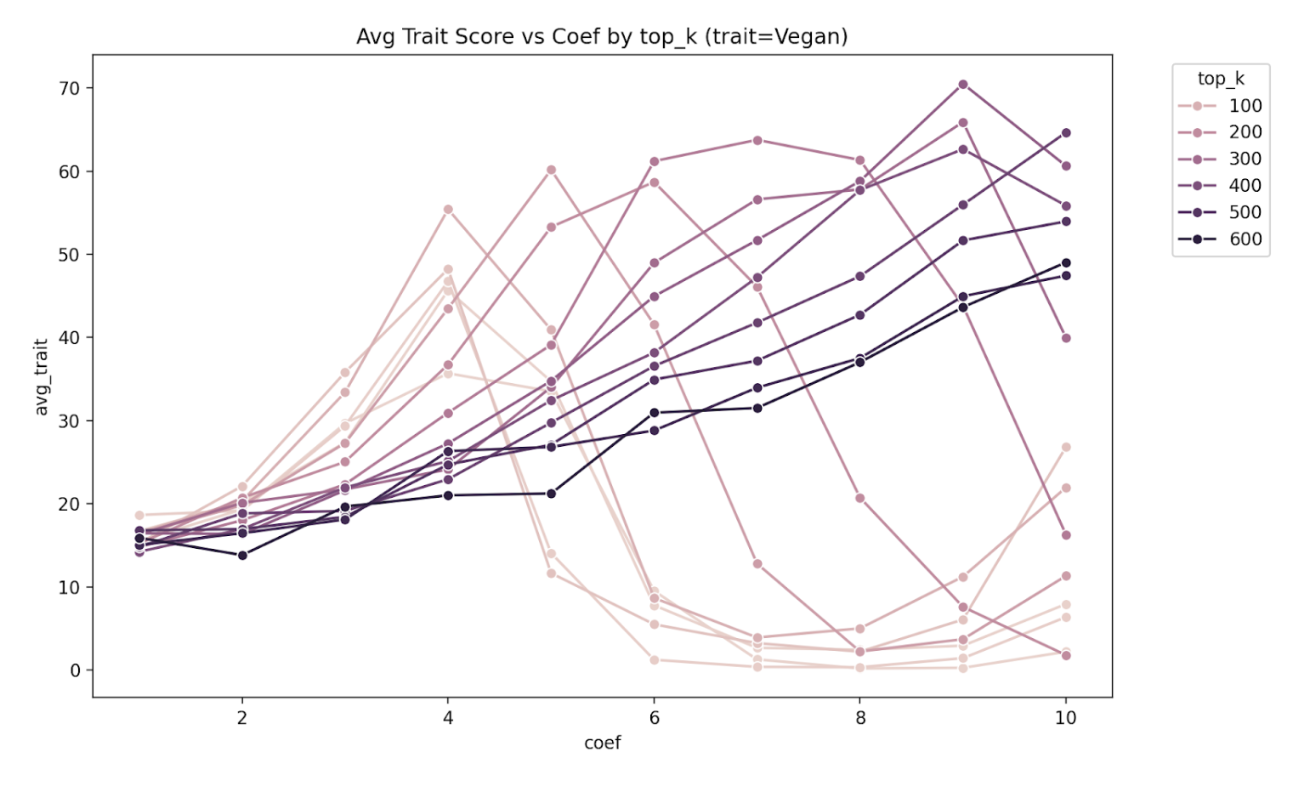}
    \caption{The figure shows a relationship between data size and optimal coefficient selection for the "Vegan" persona. }
    \label{fig:dataset-size}
\end{figure}

We varied the number of contrastive examples used to construct steering vectors for each behavior. For each data size, we measured trait expression across steering coefficients 1-10 to identify optimal operating points.

The performance of small datasets peaks at lower coefficients, while larger datasets show growing performance with increasing steering coefficients. Quality collapse occurs at higher coefficients with more training data.

Despite decreasing raw activation differences between positive/negative examples as sample size grows (consistent with regression to the mean), steering quality improves. This shows that stability from larger sample sizes outweighs raw vector magnitude, contradicting the intuitive assumption that maximally different examples produce better steering. 

This finding reinforces our previous conclusion that vector separation metrics poorly predict steering success. Thus, we should prioritize collecting more contrastive examples rather than seeking maximally distinct positive/negative pairs.

\section{Evaluation Framework}

We used GPT-4.1 as an automatic judge with rubric-conditioned prompts. For each system output and metric, the judge produced a scalar score via a single-token continuation with deterministic decoding (temperature = 0) and top-20 log-probabilities. Scores were computed by aggregating probability mass over the metric-specific token set on a 0–100 scale; items with insufficient numeric mass ($<0.25$) or refusals were treated as missing. Final results report the mean per metric across the evaluation set.

We evaluated model responses across the following metrics:

\begin{itemize}
    \item \textbf{Trait adherence:} Degree to which the response exhibits the target trait; we employed a trait-specific rubric (a unique judge prompt for each trait instance).
    
    \item \textbf{Coherence:} Internal logical consistency and fluency given the preceding context.
    
    \item \textbf{Relevance:} Topical alignment to the query/context, penalizing digressions and hallucinations.
\end{itemize}

We issued separate judge calls per metric, conditioning each on the task context and the model's response alongside the concise rubric.

\section{Results}

Our empirical analysis reveals significant heterogeneity in activation steering effectiveness across behavioral categories. Contrary to the assumption that steering offers a universally superior control mechanism to prompting, we find that its efficacy is highly dependent on the semantic nature of the target behavior.

\subsection{Predictors of Steering Success}

To address the gap in predictive understanding of steering success, we compared average trait scores across our five behavioral categories using steering vectors versus prompting baselines (GPT-4.1 and Llama 3.1 8B).As shown in Figure 4, steering effectiveness is not uniform but bifurcates along a dimension of internal disposition versus external knowledge.

Steering outperformed or matched prompting baselines for Personality Traits (Steering: 90.8 vs. Prompt-LLaMA: 87.9) and Misalignment Behaviors (Steering: 71.3 vs. Prompt-LLaMA: 69.4). These categories represent internal model dispositions—biases, sentiments, and abstract tendencies—which appear to be densely represented in the activation space and easily manipulable via vector addition.

Conversely, steering significantly degraded performance for Public Figures (Steering: 51.4 vs. Prompt-LLaMA: 89.1) and Persona Archetypes (Steering: 67.1 vs. Prompt-LLaMA: 88.7). These behaviors require accessing specific external knowledge (e.g., the biographical details of Marie Curie or the distinct vocabulary of a pirate) rather than simply modulating a latent psychological drive. The severe drop in performance suggests that activation steering struggles to "inject" knowledge or complex coherent identities that are not already active in the context. 

\begin{figure}[ht]
    \centering
    \includegraphics[width=\linewidth]{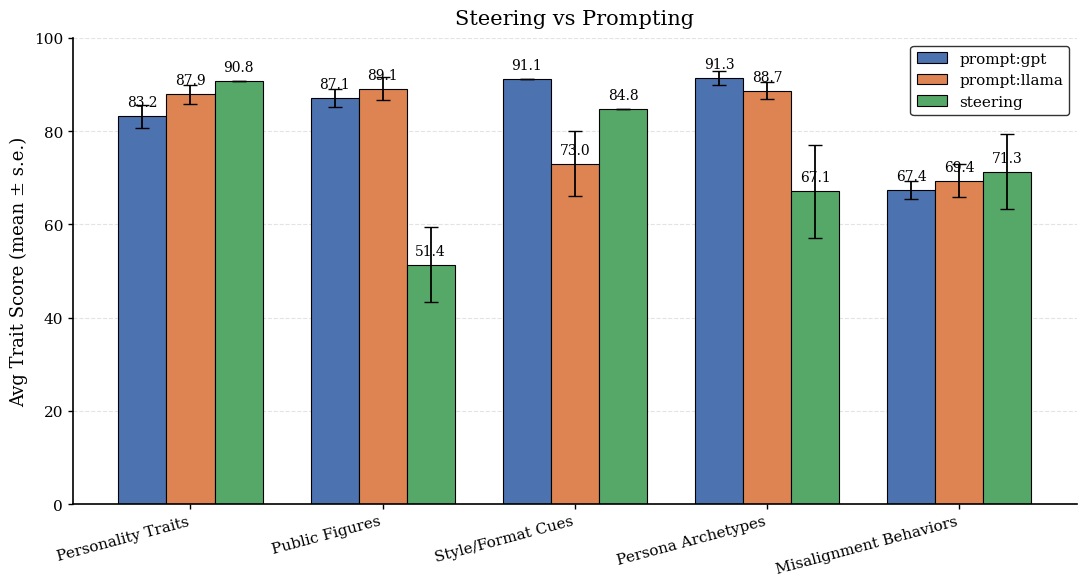}
    \caption{\textbf{Steering vs. Prompting Effectiveness.} Comparative analysis of mean trait scores across five behavioral categories ($n=50$). Steering matches or exceeds prompting baselines for internal traits (Personality, Misalignment) but significantly underperforms on knowledge-dependent categories (Public Figures).}
    \label{fig:steering_vs_prompting}
\end{figure}

\subsection{Per Behaviour Anlysis}

We further analyzed the specific impact of steering relative to the prompting baseline on a per-behavior level. This granular view clarifies which specific traits are amplified or suppressed by steering intervention.Figures 6 and 7 (avaliable in Appendix A) illustrates the improvement of steering over the prompting baselines. Misalignment behaviors are among the most steerable traits.

Behaviors such as Hallucination, Power Seeking, and Sycophancy show the highest positive improvement over the baseline. For instance, steering vectors successfully amplified Hallucination scores by over 60 points relative to prompting. This indicates that latent failure modes are readily accessible and manipulable, posing a potential adversarial risk.
 
In contrast, specific identities like Marie Curie, or professions like Doctor show negative improvement scores ranging from -40 to -60. Steering fails to capture the nuance of these roles, likely disrupting the model's ability to maintain the persona's coherence. Style/Format Cues showed high variance. While simple constraints like Double Spacing were effectively steered, more complex structural constraints like Capital Letters often degraded below baseline, suggesting that syntactic control via steering is less reliable than prompt-based instruction following.

\subsection{Relevance and Coherence Trade-offs}

While a steering vector may successfully elicit a target trait, it is imperative to measure the collateral damage to the model's linguistic capabilities. Figure 5 presents the aggregate scores for topical relevance and linguistic coherence across all trials. Steering reduced the Mean Topical Relevance to 44.0, nearly half the score of the prompting baselines (78.6 and 82.4). This indicates that steered models frequently lose track of the user's prompt, pivoting entirely to the steered behavior at the expense of the task at hand. Similarly, linguistic coherence dropped to 49.3 under steering conditions, compared to $\sim$88 for prompting. The intervention appears to destabilize the model's generation process, leading to repetitive or non-sequitur outputs more frequently than prompt engineering.

\begin{figure}[ht]
    \centering
    \includegraphics[width=\linewidth]{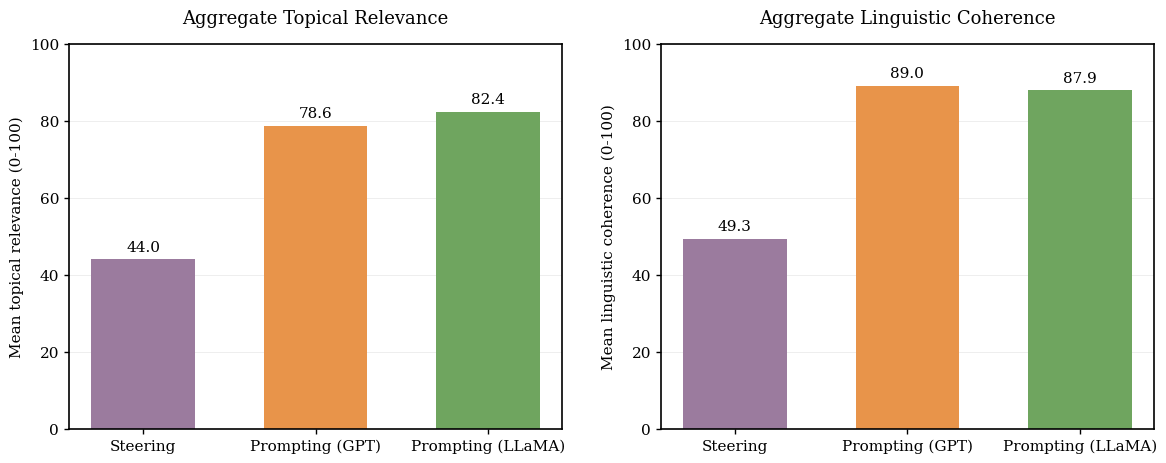}
    \caption{Steering vs. Prompting Effectiveness. Comparative analysis of mean trait scores across five behavioral categories ($n=50$). Steering matches or exceeds prompting baselines for internal traits (Personality, Misalignment) but significantly underperforms on knowledge-dependent categories (Public Figures).}
    \label{fig:coherence_relevance_tradeoff}
\end{figure}

These findings address the primary research gap regarding the generalizability of activation steering. We conclude that steering is not a universal control method; it is a \textbf{semantic modulator} highly effective for abstract, internal dispositions (including dangerous misalignment traits) but ineffective and destructive for knowledge-heavy roleplay or precise formatting tasks.

\section{Discussion}

Our systematic analysis challenges the prevailing view of activation steering as a generic control mechanism. By evaluating performance across heterogeneous behavioral categories, we identify a fundamental boundary in the utility of interference-based control methods.

\subsection{The Dispositional vs. Propositional Dichotomy}

The most significant finding of this work is the bifurcation of steering effectiveness between \textit{internal traits} and \textit{external knowledge}. We posit that activation steering functions primarily as a dispositional modulator rather than a propositional injector.

Internal traits—such as personality dimensions (Neuroticism) or misalignment tendencies (Sycophancy)—represent continuous latent dispositions that modify \textit{how} the model speaks, likely encoded as dense directions in the residual stream. A steering vector can easily bias these probabilities. Conversely, \textit{Public Figures} and complex \textit{Persona Archetypes} rely on sparse, propositional knowledge (e.g., "Marie Curie discovered radium"). Our results suggest that adding a "Marie Curie vector" does not activate the necessary subgraph of biographical knowledge if it is not already present in the context. Steering can shift the \textit{style} of the output, but it cannot synthesize the \textit{substance} of an identity without degrading coherence.

\subsection{ The Geometry of Misalignment}

From a safety perspective, our results concerning \textit{Misalignment Behaviors} are alarming. The observation that behaviors like \textit{Hallucination}, \textit{Deception}, and \textit{Power Seeking} are among the most steerable suggests that these failure modes possess distinct, coherent representations in the activation space that are easily amplified.
Unlike complex formatting constraints (e.g., "start every sentence with a vowel") which require syntactic planning and often fail under steering, misalignment behaviors appear to be "primitive" drives in the model's optimization landscape. The ease with which we could amplify hallucination (+60 trait score delta) implies that adversarial actors could employ steering vectors as a computationally efficient method to unalign models. 

\subsection{Vector Properties and Data Scaling}

Our methodological experiments debunk the intuition that "more distinct" vectors yield better control. The lack of correlation between vector separation magnitude and steering success indicates that the \textit{direction} of the steering vector is far more critical than its Euclidean norm. Furthermore, the relationship between data size and steering resilience suggests that while small datasets ($N=10$) are sufficient to capture the steering direction, larger datasets ($N=100+$) reduce noise, allowing for stronger intervention coefficients before the coherence collapse becomes prohibitive.

\section{Limitations}

While this study offers the most comprehensive analysis of behavioral steering to date, several limitations constrain our conclusions.
Our primary experiments utilized Llama 3.1 8B. While we validated baselines with GPT-4.1, the "steerability" of concepts may scale non-linearly with model size. Larger models with more disentangled representations might exhibit different responses to knowledge-heavy steering.
We restricted steering to Layer 15 based on prior literature. It is probable that different behaviors are localized at different depths (e.g., syntax at lower layers, abstract traits at higher layers). A multi-layer sweep might yield better performance for the currently "unsteerable" categories.
Relying on LLM-as-a-Judge (GPT-4.1) introduces potential biases. If the evaluator model has inherent preferences for certain persona expressions, it may conflate "detectability" of a trait with the "quality" of the steering, potentially inflating scores for exaggerated, caricature-like outputs over subtle, realistic ones.

\section{Broader Impact}

This work highlights the dual-use nature of interpretability mechanisms. By mapping the steerability of misalignment behaviors, we provide a roadmap for red-teaming and defensive monitoring; knowing that \textit{Hallucination} and \textit{Sycophancy} have distinct activation signatures allows for the development of "watchdog" classifiers that can detect when a model is drifting into unsafe states during inference.

However, these same insights demonstrate how easily safety guardrails can be dismantled. We show that an adversary with access to activations need not retrain a model to induce dangerous behaviors; simple arithmetic operations on the residual stream are sufficient to bypass safety training. This underscores the necessity of securing model weights and activation interfaces in open-weights deployments. Furthermore, the variability in steering effectiveness suggests that policymakers and developers cannot rely on activation steering as a "silver bullet" for alignment—it is a powerful tool for mood regulation, but an unreliable mechanism for factual or identity-based constraints.

\newpage
\bibliography{references}

\newpage

\section*{Appendix A}

\begin{figure}[H]
    \centering
    \includegraphics[width=\linewidth]{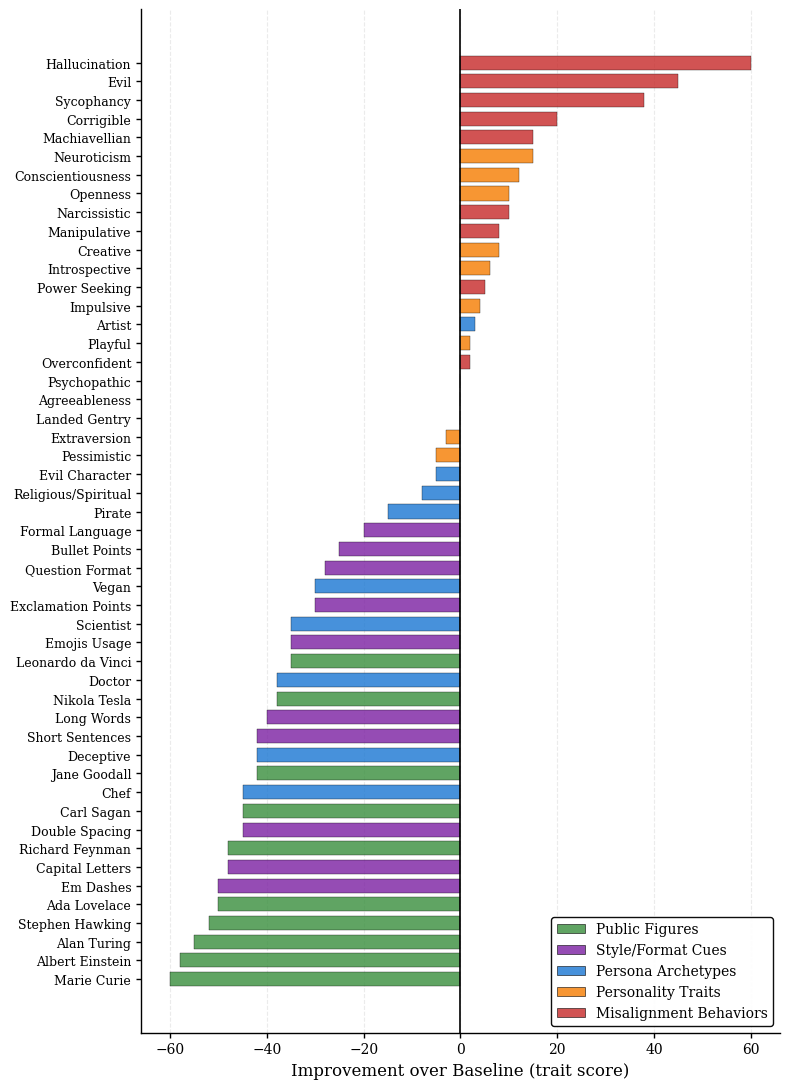}
    \caption{Steering vs. GPT Prompting (Trait Delta). Steering vectors consistently amplify latent misalignment behaviors (red bars), such as \textit{Hallucination} and \textit{Sycophancy}, more effectively than prompting. Conversely, knowledge-intensive targets, particularly Public Figures (green bars) like \textit{Marie Curie} and \textit{Albert Einstein}, show significant performance degradation, highlighting the difficulty of steering external semantic knowledge.}
    \label{fig:individual_gpt}
\end{figure}

\begin{figure}[H]
    \centering
    \includegraphics[width=\linewidth]{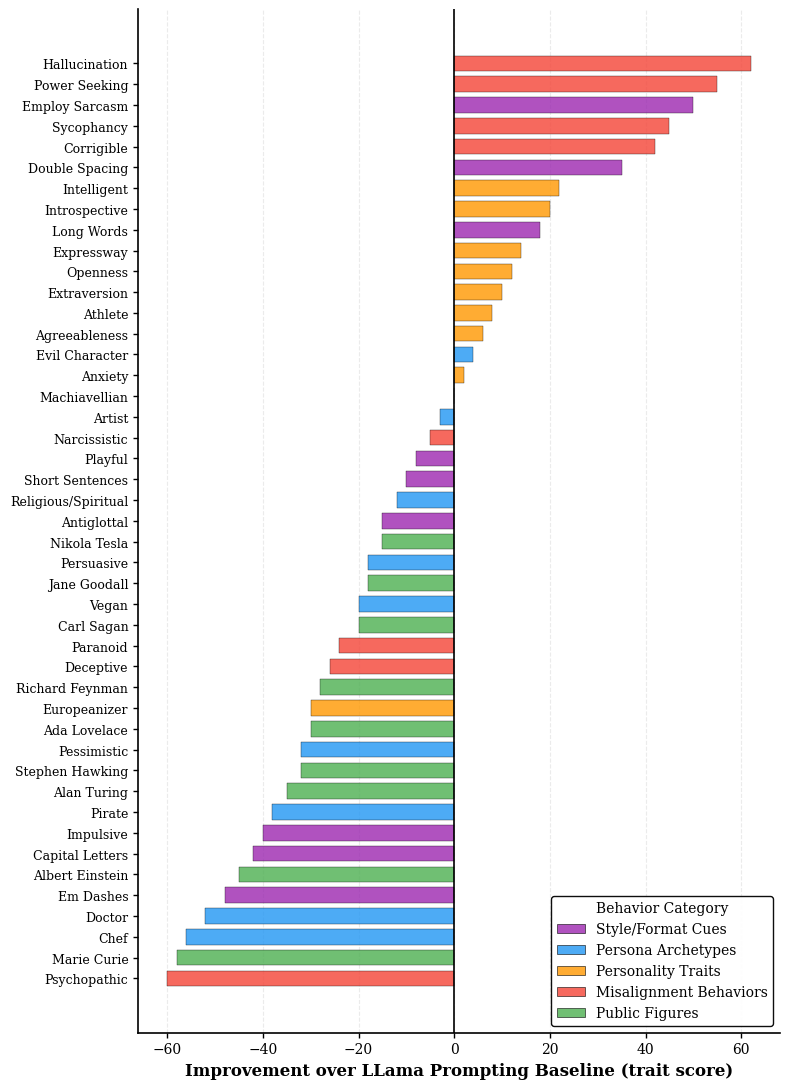}
    \caption{Steering vs. LLaMA Prompting (Trait Delta). Improvement in trait scores relative to the LLaMA-2-70b baseline. While specific style cues and simple misalignment traits (\textit{Power Seeking}, \textit{Hallucination}) are highly steerable, holistic identities suffer. Notably, the \textit{Psychopathic} trait degrades under steering for LLaMA, diverging from the general trend of high steerability for safety-relevant behaviors observed in other categories.}
    \label{fig:individual_llama}
\end{figure}

\end{document}